\documentclass{article}

\usepackage[sglblindworkshop, final]{ai4nextg_neurips_2025}

\usepackage[utf8]{inputenc} 
\usepackage[T1]{fontenc}    
\usepackage{hyperref}       
\usepackage{url}            
\usepackage{booktabs}       
\usepackage{amsfonts}       
\usepackage{nicefrac}       
\usepackage{microtype}      
\usepackage{xcolor}         

\usepackage{graphicx}
\usepackage{array}
\usepackage{amsmath}
\usepackage{multirow} 
\usepackage{enumitem}
\usepackage{algorithm}
\usepackage[noend]{algpseudocode}

\title{Dynamic Features Adaptation in Networking: \\ Toward Flexible training and Explainable inference}

\author{%
  Yannis Belkhiter\\
  IBM Research Europe\\
  Trinity College Dublin\\
  \texttt{yannis.belkhiter@ibm.com} \\
  \And
  Seshu Tirupathi\\
  IBM Research Europe\\
  Dublin, Ireland\\
  \texttt{seshutir@ie.ibm.com} \\
  \And
  Giulio Zizzo\\
  IBM Research Europe\\
  Dublin, Ireland\\
  \texttt{giulio.zizzo2@ibm.com} \\
  \AND
  Merim Dzaferagic\\
  Trinity College Dublin\\
  Dublin, Ireland\\
  \texttt{merim.dzaferagic@tcd.ie} \\
  \And
  John D. Kelleher\\
  ADAPT Research Centre\\
  Trinity College Dublin\\
  \texttt{john.kelleher@tcd.ie} \\
}

\begin{document}

\maketitle

\begin{abstract}
  As AI becomes a native component of 6G network control, AI models must adapt to continuously changing conditions, including the introduction of new features and measurements driven by multi-vendor deployments, hardware upgrades, and evolving service requirements. To address this growing need for flexible learning in non-stationary environments, this vision paper highlights Adaptive Random Forests (ARFs) as a reliable solution for dynamic feature adaptation in communication network scenarios. We show that iterative training of ARFs can effectively lead to stable predictions, with accuracy improving over time as more features are added. In addition, we highlight the importance of explainability in AI-driven networks, proposing Drift-Aware Feature Importance (DAFI) as an efficient XAI feature importance (FI) method. 
  DAFI uses a distributional drift detector to signal when to apply computationally intensive FI methods instead of lighter alternatives.
  Our tests on 3 different datasets indicate that our approach reduces runtime by up to 2 times, while producing more consistent feature importance values. Together, ARFs and DAFI provide a promising framework to build flexible AI methods adapted to 6G network use-cases.
\end{abstract}

\section{Introduction and Related Work}
Future communication networks are envisioned as AI-native, where artificial intelligence is integrated into all layers of the network. Two complementary perspectives dominate this vision: AI for 6G, where AI algorithms drive the control and optimization of the network, and 6G for AI, where next-generation networks provide the substrate for AI services \cite{dzaferagic2024decentralized}. In this paper, we focus on AI for 6G. We study how AI can be embedded into heterogeneous base stations and network nodes to identify mobility patterns based on Radio Access Network (RAN) Key Performance Indicators (KPIs).

The deployment of AI in live networks presents distinct challenges. First, AI control is inherently data-driven, requiring access to KPIs and measurements across different nodes. However, due to vendor-specific implementations, hardware generations, and configuration settings, not all network nodes expose the same set of features \cite{mestoukirdi2025reliable,trichias2025ai}. This heterogeneity makes it difficult to train and deploy a single global model across the network. Second, even if all features were consistently available, local conditions vary greatly. A base station at a highway may mostly serve vehicular users, while one in a city center may see a mix of pedestrians, buses, and static users. These environmental differences imply that local fine-tuning is critical \cite{samsung2025ainative6g}. Third, data drift is inevitable: traffic patterns, user behavior, and radio conditions evolve over time, rendering static models ineffective \cite{qualcomm2024adaptive6g}. Finally, as AI takes on an increasingly central role in controlling critical infrastructure, explainability becomes essential to ensure trust, operator acceptance, and regulatory compliance.

Feature Importance (FI) offers a promising way of addressing these challenges. In general, FI methods refer to techniques that assign scores to input features based on their contribution to the predictions of a model \cite{molnar2018interpretable,rajbahadur2022ImpactofFeatureImportance}. These algorithms assign higher scores to features that significantly influence predictions, helping the user to identify which variables most strongly affect the output. For both the effectiveness of the training pipeline and interpretability, FI is a crucial piece of information that enables better comprehension of the prediction and optimization of machine learning models \cite{molnar2018interpretable,scornet2021treesforestsimpuritybasedvariable}, echoing challenges seen in the Network space. In this context, our main contributions include: 
\begin{itemize}[leftmargin=*, itemsep=0pt]
    \item \textbf{Effectiveness and adaptation of Adaptive Random Forests (ARFs) on network datasets.}
We show that applying ARFs to communication datasets with iterative training provides strong predictive performance. Our results confirm that progressively increasing the number of available features leads to higher accuracy, illustrating the value of applying ARFs network environments. 
    \item \textbf{DAFI as a promising explainable AI approach for networks.}
We present \emph{Drift-Aware Feature Importance} (DAFI), an algorithm that uses distributional drift to decide when to apply costly but accurate explainability methods versus lightweight alternatives. This drift-aware strategy leads to more efficient iterative monitoring and training of ARFs. In doing so, DAFI provides a practical way to couple explainability with model adaptability in network contexts.
\end{itemize}

\section{Limitation of current XAI approaches}


The current state-of-the-art of FI methods falls into two categories: model agnostic and model specific \cite{rajbahadur2022ImpactofFeatureImportance}. 
Our approach combines model agnostic (SHAP) and model specific (Mean Decrease in Impurity (MDI)) methods (see Appendices \ref{sec:shap-theory} and \ref{sec:MDI-method}, respectively). 
We selected SHAP and MDI as representative baselines because they are widely adopted state-of-the-art methods \cite{molnar2018interpretable}, including in the Networking field \cite{brik2024explainableai6goran}. Although incremental XAI methods have obtained promising results in streaming scenarios \cite{Fumagalli_2023, Muschalik_2022}, their adoption is still limited. In the Networking space, SHAP (or variants) has been applied for KPI reduction in O-RAN controllers \cite{Tassie2024_SHAP_KPI} and traffic classification tasks \cite{Nascita_2023_traffic_SHAP}, while MDI is a common method for tree-based models such as Adaptive Random Forests.

\textbf{The need for more efficient feature importance predictors:} Although SHAP is the most widely used method for feature importance for its accurate FI predictions \cite{rajbahadur2022ImpactofFeatureImportance}, it is computationally expensive \cite{molnar2018interpretable}. In the context of communication networks, this limitation becomes critical: base stations operate continuously and in real-time. It means that FI values must be recomputed at short intervals of time to capture sudden changes from the available features (i.e. KPIs). Relying solely on SHAP can harm the runtime \cite{basaran2025xainomalyexplainableinterpretabledeep}. Appendices \ref{sec:appendix_limit-shap} and \ref{sec:shap-runtime-limit} further supports this claim, showing the exponential growth of the SHAP runtime when adding new features or increasing the size of the model, respectively. 
\\ %
In comparison to SHAP, MDI is a common tree-based model specific FI metric, and is a much faster method. However, MDI often disagrees with SHAP analysis, especially when the model is not well adapted to the current data as MDI relies on a model's performance \cite{molnar2018interpretable,rajbahadur2022ImpactofFeatureImportance,Saarela2021}.


\section{DAFI: Drift-aware feature importance}

Both model agnostic and model specific XAI FI methods come with limitations. Although SHAP analysis offers the best feature importance score based on the complex and evolving data stream, this method is expensive to run. On the other hand, MDI values are quick to compute but are often not aligned with the SHAP values, especially when there is drift in the data distribution and the tree-based model has not yet been updated accordingly.

\begin{figure}[h]
    \centering
    \vspace{-0.25cm}
    \includegraphics[width=0.9\linewidth]{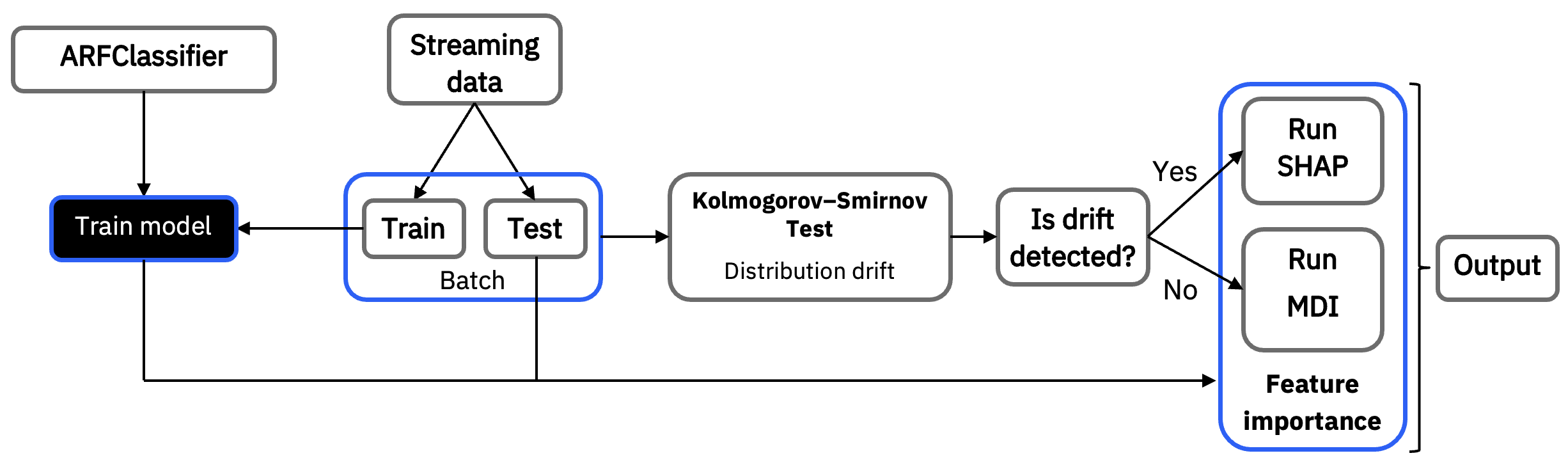}
    \vspace{-0.3cm}
    \caption{DAFI: Drift-aware Feature Importance algorithm}
    \label{fig:drift-aware-algorithm}
\end{figure}

The central challenge we address in this work is determining when to use accurate but costly SHAP values versus efficient MDI values, without compromising on the reliability of the results. To achieve this, we propose DAFI (see Figure \ref{fig:drift-aware-algorithm}), a \emph{Drift-Aware feature importance} algorithm which dynamically selects the appropriate method based on the observed data distribution shifts. Since the performance of the ARF is subject to drifts in the data distribution, our approach employs SHAP values in the presence of drift and defaults to MDI otherwise. Drift is assessed with the \emph{Kolmogorov-Smirnov (KS) Test} \cite{KS_test} (see Appendix \ref{sec:ks_test_theory}). This combination of a drift detection technique with a decision-making framework yields an adaptive, efficient FI approach for dynamic data streams.

\textbf{DAFI Algorithm.} We build our DAFI algorithm upon the KS-Test drift detection metric that we presented. For each sample $\mathbf{x}$, DAFI computes the SHAP values if a drift is detected, otherwise it computes the MDI. For each feature, the drift is computed between the previous and current train batches, assuming that we do not have access to the testing data yet. Thus the drift detection can assess if the model has required significant changes in the current training batch. The user needs to set a threshold $\eta$ for the KS-Test. Experimentally, we have set $\eta = 1$ for the Electricity and Weather datasets, and $\eta = 0.125$ for the Network dataset since it has much larger batches, so drifts are less pronounced. Appendix \ref{sec:appendix-DAFI-algorithm} formally defines our DAFI algorithm.

\vspace{-0.2cm}

\section{Experimental Setup} \label{sec:experimental-step}

\vspace{-0.25cm}


\textbf{Models and datasets.} To test our novel algorithm, we use the \emph{Adaptive Random Forest Classifier} from the river library \cite{montiel2020rivermachinelearningstreaming} and 3 binary classification datasets from diverse domains (Networks, Electricity, and Weather). Appendix \ref{sec:dataset-description} describes each dataset. Our main focus is on the Network use-case, so in the main text we only present the results obtained on the \emph{Network} dataset, and include results from other datasets in Appendix \ref{sec:appendix-evaluation-other-datasets}.

\textbf{Iterative training setup.} For each evaluation, we introduce an iterative training setup. We conducted a prequential split of the datasets in equal batches (80:20 train test - keeping the order). We created 50 batches for the Electricity and Weather datasets, and 40 batches for Network. The reduced number of batches for the Network dataset was chosen for two  reasons. First, due to the strong class imbalance in the Network dataset, smaller batches occasionally contained only a single label, limiting the complexity of the stream. Second, using fewer batches allows us to demonstrate that the algorithm remains relatively stable across different batch sizes, as long as both classes are present. 

\textbf{Dynamic feature adaptation.} To assess the performances of both model and FI methods in a complex and dynamic data stream, we varied the number of features available as training progressed. For each dataset, we started the training with 3 features, added one more feature at epoch 10, and two more features at epoch 20. We ran Feature Importance methods on $n_\text{samples}$ samples randomly picked in the testing dataset. For all the experiments, we set our models with $n_\text{trees} = 10$, and $n_\text{samples} = 50$. All our experiments were carried out on an Apple Macbook Pro M3, 24GB RAM, with a seed of $42$.

\textbf{Metrics.} To assess the adaptability of the ARFs on the Network dataset, we report the \emph{training} and \emph{testing accuracies}. To test the efficiency of our DAFI algorithm, we rely on two criteria. First, the \emph{runtime} of feature importance computation is our first way of assessing our algorithm. As SHAP is computationally expensive, one of the objectives of our algorithm is to decrease the runtime. Second, the \emph{feature importance accuracy} of MDI is assessed against SHAP. As SHAP is a reliable and consistent way of computing feature importance, it is used as a ground-truth for our experiment. 

\textbf{Top-k features} In our complex stream configuration, setting a fixed and reliable $k$ is challenging. Naively, a simple Top-$k$ evaluation selects the $k$ most important features based on their ranked importance scores and compares them between the FI method and SHAP. However, as we are dealing with normalized scores, feature importance scores depend on the number of features present in the dataset. To address this, we introduce a \emph{Dynamic Top-$k$} evaluation, which adapts to variations in feature importance weights. Instead of a fixed Top-$k$ metric, it determines the optimal number of top features to consider by accumulating SHAP feature importance weights in descending order until a predefined cumulative threshold is reached. For our experiment, we fixed this threshold to $0.8$. Based on the selected Top-$k$ features given by SHAP, we are assessing performance of MDI using \emph{Set} and \emph{Exact matches} analysis. Set matches returns 1 when Top-$k$ of SHAP and MDI sets are same (irrespective of the order or sequence of features), and exact matches returns 1 when Top-$k$ ranks are equal, i.e. feature sequence is also the same. Otherwise, the metrics return 0 (see Appendix \ref{sec:appendix-topk-illustration}). 

\textbf{Spearman's rank.} To further analyze the effectiveness of our algorithm, we evaluate the correlation between the feature rankings produced by MDI and those generated by SHAP \cite{rajbahadur2022ImpactofFeatureImportance}. Specifically, we use \emph{Spearman's rank correlation coefficient}, normalized between 0 and 1. As it is often used in ranking classification contexts, this additional metric helps us to consider more flexibly the differences in the entire ranking of feature importance.

\section{Evaluation}

\begin{figure}[h]
    \centering
    \includegraphics[width=0.95\linewidth]{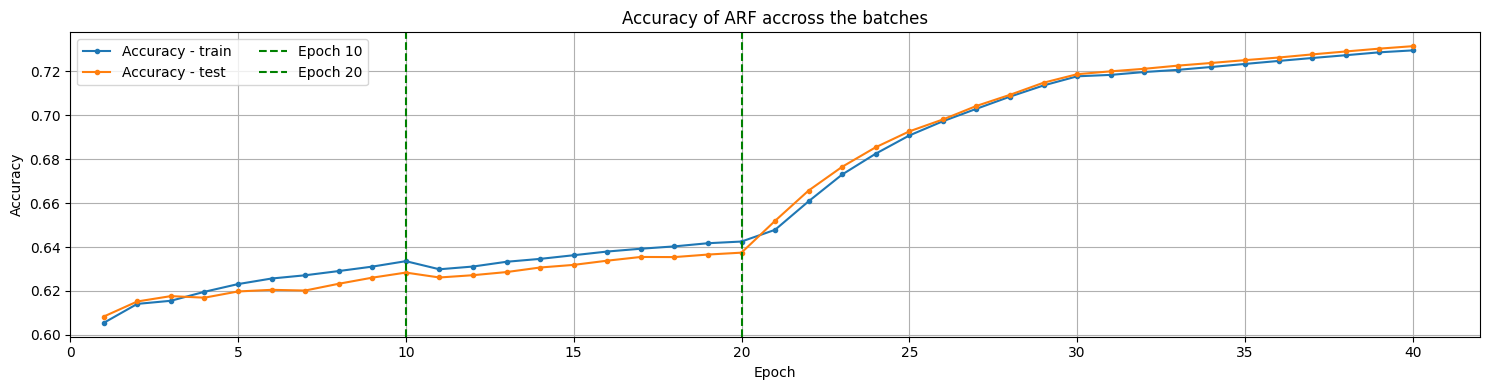}
    \vspace{-0.3cm}
    \caption{\small Performance of ARF for benign and attack network classifications with evolving Network KPIs - Network dataset \cite{Xavier2024} - epoch 0: start with features "mac\_dl\_mcs", "phy\_ul\_pusch\_sinr", "phy\_ul\_pucch\_sinr" - epoch 10: adding "phy\_ul\_pusch\_rssi" - epoch 20: adding "mac\_dl\_cqi", "phy\_ul\_pucch\_rssi" - $n_{\text{model}}=10$}
    \label{fig:arf-networking}
\end{figure}

\textbf{Performance of ARFs on Evolving Network Features.} 
Figure \ref{fig:arf-networking} illustrates the performance of ARFs on the Network dataset under evolving feature sets, where new KPIs are progressively introduced at epoch 10 and 20 (1 and 2 features added, respectively). We observe that the addition of new features does not harm, but increases the model's performance. Specifically, the introduction of two new features at epoch 20 improves the accuracy of the ARF from approximately $0.64$ to $0.72$. The results highlight the stability of the ARF model dynamic feature spaces, 
such as those that may arise due to differences in vendor-specific implementations, or configuration settings in communication networks.


\textbf{Drif-Aware Feature Importance (DAFI) for Efficient Explainability.} We present the results of our DAFI algorithm, where accuracy metrics are computed element-wise, and then averaged over the $n=50$ samples. Table \ref{fig:tab-element-wise} presents for each FI method, on the Network dataset, the runtime along with the discussed FI metrics. Appendix \ref{sec:appendix-evaluation-other-datasets} reports the results on the Electricity and Weather datasets.

\begin{table}[h]
\centering
\footnotesize
\begin{tabular}{p{1.2cm}>{\centering\arraybackslash}p{1.6cm}>
{\centering\arraybackslash}p{1.7cm}>{\centering\arraybackslash}p{1.4cm}>{\centering\arraybackslash}p{1.4cm}>{\centering\arraybackslash}p{1.6cm}>{\centering\arraybackslash}p{2.4cm}}
\toprule
\textbf{Datasets} & \textbf{FI Method} & \textbf{Runtime (s)} & \textbf{Saved (\%)} & \textbf{Top-k set} & \textbf{Top-k exact} & \textbf{Top-k Spearman} \\
\midrule
\multirow{3}{*}{Network} & SHAP & 1488.32 & 0.00 & 1.00 & 1.00 & 1.00 \\
& MDI & 9.37 & 99.37 & 0.27 & 0.09 & 0.53 \\ 
& \textbf{DAFI} & 674.49 & 54.68 & 0.55 & 0.40 & 0.67 \\ 
\bottomrule
\end{tabular}
\caption{Runtime and metrics of FI methods applied Element-wise - average over the batches}
\label{fig:tab-element-wise}
\end{table}

Regarding the Runtime, SHAP consistently has the highest runtime across each dataset. In contrast, MDI is orders of magnitude faster compared to SHAP, with a consistent saved runtime score of more than 99\%. Drift-Aware achieves a good balance between SHAP and MDI, substantially reducing the computational cost. The saved runtime obtained by our algorithm is notably high, reaching 55\% for the Network dataset and around 35\% for the Electricity and Weather datasets. 

\noindent However, MDI obtained a relatively low Top-$k$ exact match score on average, confirming our observations from the literature.  The DAFI algorithm outperformed MDI for all metrics, demonstrating substantial improvements in Top-$k$ exact match (up to 0.76 in Electricity) and Spearman Top-$k$ match (0.85 in Electricity and 0.83 in Weather) - where MDI reported low scores. This suggests that our algorithm improved the consistency of FI prediction, effectively switching between SHAP and MDI.

\section{Conclusion}
We demonstrated that ARFs, when trained iteratively, can effectively capture the dynamics of communication data streams, with accuracy improving as more features are introduced. To complement this, we proposed DAFI, a drift-aware feature importance method that reduces computational overhead while guiding the selection of the most relevant features. Together, ARFs and DAFI highlight a promising direction for AI-native 6G networks: models that adapt to evolving feature spaces, remain resilient to distributional shifts, and provide transparent insights into their decisions.


\textbf{Future work.} While these results highlight the method's potential as a practical alternative to computationally expansive approaches, future research should focus on further validating this method across additional datasets and adaptive scenarios (such as feature removal), as well as exploring its integration with other types of ensemble models, and other FI methods. 



\section*{Acknowledgment}

This work has been partially supported by the 6G-XCEL project (grant agreement 101139194), funded by the EU Horizon Europe program. 

\bibliography{main}

\begin{thebibliography}{10}

\bibitem{dzaferagic2024decentralized}
Merim Dzaferagic, Marco Ruffini, Nina Slamnik-Krijestorac, Joao~F Santos, Johann Marquez-Barja, Christos Tranoris, Spyros Denazis, Thomas Kyriakakis, Panagiotis Karafotis, Luiz DaSilva, et~al.
\newblock Decentralized multi-party multi-network ai for global deployment of 6g wireless systems.
\newblock {\em arXiv preprint arXiv:2407.01544}, 2024.

\bibitem{mestoukirdi2025reliable}
Mohamad Mestoukirdi and Mourad Khanfouci.
\newblock Reliable vertical federated learning in 5g core network architecture.
\newblock {\em arXiv preprint arXiv:2505.15244}, 2025.

\bibitem{trichias2025ai}
K~Trichias, M~Christopoulou, P~Porambage, C~Benzaid, V~Vassiliou, D~Artu{\~n}edo~Guill{\'e}n, D~Tsolkas, Marios Avgeris, C~Papagianni, A~Jain, et~al.
\newblock Ai/ml as a key enabler of 6g networks: Methodology, approach and ai-mechanisms in sns ju.
\newblock 2025.

\bibitem{samsung2025ainative6g}
{Samsung Research}.
\newblock Ai-native and sustainable communication: Samsung 6g white paper.
\newblock Technical report, Samsung Electronics, June 2025.
\newblock Accessed: 2025-08-27.

\bibitem{qualcomm2024adaptive6g}
{Qualcomm Technologies, Inc.}
\newblock How will adaptive intelligence in 6g transform wireless connectivity: 6g foundry white paper.
\newblock Technical report, Qualcomm, 2024.
\newblock Accessed: 2025-08-27.

\bibitem{molnar2018interpretable}
Christoph Molnar.
\newblock {\em Interpretable Machine Learning}.
\newblock 3 edition, 2025.

\bibitem{rajbahadur2022ImpactofFeatureImportance}
Gopi~Krishnan Rajbahadur, Shaowei Wang, Gustavo~A. Oliva, Yasutaka Kamei, and Ahmed~E. Hassan.
\newblock The impact of feature importance methods on the interpretation of defect classifiers.
\newblock {\em IEEE Transactions on Software Engineering}, 48(7):2245–2261, July 2022.

\bibitem{scornet2021treesforestsimpuritybasedvariable}
Erwan Scornet.
\newblock Trees, forests, and impurity-based variable importance, 2021.

\bibitem{brik2024explainableai6goran}
Bouziane Brik, Hatim Chergui, Lanfranco Zanzi, Francesco Devoti, Adlen Ksentini, Muhammad~Shuaib Siddiqui, Xavier Costa-Pérez, and Christos Verikoukis.
\newblock Explainable ai in 6g o-ran: A tutorial and survey on architecture, use cases, challenges, and future research, 2024.

\bibitem{Fumagalli_2023}
Fabian Fumagalli, Maximilian Muschalik, Eyke Hüllermeier, and Barbara Hammer.
\newblock Incremental permutation feature importance (ipfi): towards online explanations on data streams.
\newblock {\em Machine Learning}, 112(12):4863–4903, September 2023.

\bibitem{Muschalik_2022}
Maximilian Muschalik, Fabian Fumagalli, Barbara Hammer, and Eyke Hüllermeier.
\newblock Agnostic {Explanation} of {Model} {Change} based on {Feature} {Importance}.
\newblock {\em KI - Künstliche Intelligenz}, 36(3):211--224, December 2022.

\bibitem{Tassie2024_SHAP_KPI}
Chinenye Tassie, Brian Kim, Joshua Groen, Mauro Belgiovine, and Kaushik~R. Chowdhury.
\newblock Leveraging explainable ai for reducing queries of performance indicators in open ran.
\newblock In {\em ICC 2024 - IEEE International Conference on Communications}, pages 5413--5418, 2024.

\bibitem{Nascita_2023_traffic_SHAP}
Alfredo Nascita, Francesco Cerasuolo, Giuseppe Aceto, Domenico Ciuonzo, Valerio Persico, and Antonio Pescap\'{e}.
\newblock Explainable mobile traffic classification: the case of incremental learning.
\newblock In {\em Proceedings of the 2023 on Explainable and Safety Bounded, Fidelitous, Machine Learning for Networking}, SAFE '23, page 25–31, New York, NY, USA, 2023. Association for Computing Machinery.

\bibitem{basaran2025xainomalyexplainableinterpretabledeep}
Osman~Tugay Basaran and Falko Dressler.
\newblock Xainomaly: Explainable and interpretable deep contractive autoencoder for o-ran traffic anomaly detection, 2025.

\bibitem{Saarela2021}
Mirka Saarela and Susanne Jauhiainen.
\newblock Comparison of feature importance measures as explanations for classification models.
\newblock {\em SN Applied Sciences}, 3(2):272, 2021.

\bibitem{KS_test}
Frank~J. Massey.
\newblock The kolmogorov-smirnov test for goodness of fit.
\newblock {\em Journal of the American Statistical Association}, 46(253):68--78, 1951.

\bibitem{montiel2020rivermachinelearningstreaming}
Jacob Montiel, Max Halford, Saulo~Martiello Mastelini, Geoffrey Bolmier, Raphael Sourty, Robin Vaysse, Adil Zouitine, Heitor~Murilo Gomes, Jesse Read, Talel Abdessalem, and Albert Bifet.
\newblock River: machine learning for streaming data in python, 2020.

\bibitem{Xavier2024}
Bruno Xavier, Merim Dzaferagic, Marco Ruffini, and Magnos Martinello.
\newblock {RAN Performance Measurements for Security Threats}, 2024.

\bibitem{lundberg2017unifiedapproachinterpretingmodel}
Scott Lundberg and Su-In Lee.
\newblock A unified approach to interpreting model predictions, 2017.

\bibitem{kaggle_weather}
Lauri and Wu.
\newblock Will it rain tomorrow? eda and classification, 2020.
\newblock Accessed: 2025-08-15.

\bibitem{gomes_adaptive_2017}
Heitor~M. Gomes, Albert Bifet, Jesse Read, Jean~Paul Barddal, Fabrício Enembreck, Bernhard Pfharinger, Geoff Holmes, and Talel Abdessalem.
\newblock Adaptive random forests for evolving data stream classification.
\newblock {\em Machine Learning}, 106(9):1469--1495, October 2017.

\bibitem{harries1999splice2}
M.~Harries.
\newblock Splice-2 comparative evaluation: Electricity pricing.
\newblock Technical report, The University of South Wales, 1999.

\bibitem{Gama2004}
Jo{\~{a}}o Gama, Pedro Medas, Gladys Castillo, and Pedro Rodrigues.
\newblock Learning with drift detection.
\newblock In {\em SBIA Brazilian Symposium on Artificial Intelligence}, pages 286--295, 2004.

\bibitem{Raksha2021}
S~Raksha, Jasmine~S Graceline, Jani Anbarasi, M~Prasanna, and S~Kamaleshkumar.
\newblock Weather forecasting framework for time series data using intelligent learning models.
\newblock In {\em 2021 5th International Conference on Electrical, Electronics, Communication, Computer Technologies and Optimization Techniques (ICEECCOT)}, pages 783--787, 2021.

\bibitem{XAVIER2024110710}
Bruno~Missi Xavier, Merim Dzaferagic, Magnos Martinello, and Marco Ruffini.
\newblock Performance measurement dataset for open ran with user mobility and security threats.
\newblock {\em Computer Networks}, 253:110710, 2024.

\end{thebibliography}
\bibliographystyle{unsrt}


\newpage

\appendix

\section{Appendix}

\subsection{SHAP: Definition and Limitations} \label{sec:shap-theory}

\emph{SHapley Additive exPlanations} (SHAP) is a model agnostic framework for interpreting model predictions by assigning feature importance values based on cooperative game theory \cite{lundberg2017unifiedapproachinterpretingmodel}. For a given data point $\mathbf{x} = \{x_1,\dots, x_j,\dots, x_d\}$ and a model $f$, SHAP computes the contribution of each feature $x_j \in \mathbf{x}$ to the prediction $f(\mathbf{x})$ by considering all possible feature combinations. The SHAP value $\phi_j$ for feature $j$ is defined as:
\begin{equation}\label{eq:shap}
    \phi_j = \sum_{S \subseteq \{1, \dots, d\} \setminus \{j\}} \frac{|S|!(d - |S| - 1)!}{d!} \left[ f(S \cup \{j\}) - f(S) \right]
\end{equation}
where $d$ is the number of features, $S$ is a subset of features, $f(S)$ represents the model output with features in $S$ included, and $\phi_j$ is the SHAP value for feature $j$. SHAP not only indicates feature importance score, but also provides directional information - positive values increase the prediction, while negative values decrease it. However, since our focus is on overall feature contribution rather than direction, we take the absolute values of the SHAP scores, and normalize them across all features of the data. This enables score comparison with the Mean Decrease in Impurity method, that consists of positive, normalized values.

\subsection{Mean Decrease in Impurity: A model specific approach}\label{sec:MDI-method}

Path-based FI methods are model specific techniques adapted to tree-based architectures, such as Adaptive Random Forests. These algorithms offer a lightweight yet effective approach for analyzing feature importance. Unlike model agnostic methods, which analyze aggregated statistics across all subsets of features, path-based methods compute feature importance dynamically by traversing the decision paths for individual predictions \cite{scornet2021treesforestsimpuritybasedvariable}. This approach is particularly useful for adaptive scenarios, where data distribution and model parameters evolve over time. For every feature node encountered along the path, the algorithm assigns an importance weight based on the \emph{Mean Decrease in Impurity} (MDI), a measure derived from the change in node impurity.

\noindent Formalized by \cite{scornet2021treesforestsimpuritybasedvariable}, the algorithm begins by traversing the decision paths for a single  instance $\mathbf{x} = \{x_1,\dots, x_j,\dots, x_d\}$ within each tree of an ensemble model. Let $\mathcal{T} = \{T_1, T_2, \dots, T_m\}$ represent the set of $m$ trees in the ensemble model, and $T_i$ denote the $i$-th tree. For each tree $T_i$, we traverse its decision path for $\mathbf{x}$. For every decision node $n_j$ in the path, where the feature $j$ is the splitting criteria, the algorithm evaluates the contribution of the feature $j$ to the final feature importance $\varphi_{j}$:
\begin{equation}\label{eq:mdi}
\varphi_{j} = \sum_{T_i \in \mathcal{T}} \sum_{n_j \in T_i} \text{MDI}(n_j, T_i)
\end{equation}
\[
\text{where} \hspace{0.1cm}
\text{MDI}(n_j, T_i) = \text{G}_{n_j} - \frac{\text{freq}_{\text{left}}}{\text{freq}_{n_j}} \cdot \text{G}_{\text{left}} - \frac{\text{freq}_{\text{right}}}{\text{freq}_{n_j}} \cdot \text{G}_{\text{right}}
\]
\[
\text{with } \text{G}_{n} = 1 - \sum_{k=1}^{C} p_k^2 \quad \text{ and } \quad p_k = \frac{\text{freq}_k}{\text{freq}_n}
\]
$\varphi_{j}$ is the feature importance for feature $j$, $\text{MDI}(n_j, T_i)$ is the Mean Decrease in Impurity for the node $n_j$ in the tree $T_i$, $\text{G}_{n}$ is the Gini index of the node $n$, $\text{G}_{\text{left}}$ and $\text{G}_{\text{right}}$ are respectively the Gini indexes of the left and right child of $n_j$, $p_k$ is the proportion of samples in class $k$ in node $n$, $\text{freq}_k$ is the number of samples in class $k$, and $\text{freq}_n$ is the total number of samples in node $n$. After computation, the importance for a feature $\varphi_{j}$ across all nodes and trees is normalized, and features are ranked based on their normalized importance values. The algorithm is presented in appendix (Section \ref{sec:MDI-algorithm}).

\noindent While the formulation of MDI above applies to binary decision trees, it can be extended to non-binary trees by generalizing the impurity reduction term. Instead of considering only left and right children, the decrease in impurity is computed by subtracting the weighted impurities of all child nodes from the impurity of the parent node. This allows the MDI measure to be applicable to trees with any type of branching structure as defined by \cite{scornet2021treesforestsimpuritybasedvariable}.

\subsection{Limitation of SHAP} \label{sec:appendix_limit-shap}

This sub-section offers evidence that relying solely on SHAP can significantly impact the runtime.

\begin{table}[h]
\centering
\footnotesize
\begin{tabular}{p{2.3cm}>{\centering\arraybackslash}p{2.2cm}>{\centering\arraybackslash}p{2cm}>{\centering\arraybackslash}p{2.2cm}>{\centering\arraybackslash}p{2.2cm}}
\toprule
\textbf{\# of features} & \textbf{Training Runtime (s)} & \textbf{SHAP Runtime (s)} & \textbf{Training Accuracy} & \textbf{Testing Accuracy} \\
\midrule
\textbf{All (8)}        & 1074.55  & 2982  & 89.51  & 82.14 \\
\textbf{Top 7 feat. (7)}   & 1084.25 & 1414 & 89.70  & 81.40 \\
\textbf{Top 4 feat. (4)}   & 744.74 & 148.3 & 86.34 & 79.31 \\
\textbf{Top 3 feat. (3)}   & 682.88  & 61.04  & 86.04  & 77.22 \\
\textbf{Top 2 feat. (2)}   & 647.20 & 20.9  & 84.03  & 76.28 \\
\bottomrule
\end{tabular}
\caption{Comparison of ARF Performance on the Electricity dataset ~\cite{kaggle_weather} with different set of features - 80:20 train/test prequential split - SHAP runtime (including setting the explainer and inference time) for $n_\text{samples} = 1000$ - $n_\text{trees} = 100$}
\label{tab:shap-runtime}
\end{table}

\begin{table}[h]
\centering
\footnotesize
\begin{tabular}{p{2.3cm}>{\centering\arraybackslash}p{2.2cm}>{\centering\arraybackslash}p{2cm}>{\centering\arraybackslash}p{2.2cm}>{\centering\arraybackslash}p{2.2cm}}
\toprule
\textbf{\# of features} & \textbf{Training Runtime (s)} & \textbf{SHAP Runtime (s)} & \textbf{Training Accuracy} & \textbf{Testing Accuracy} \\
\midrule
\textbf{All (6)}           & 1550.4 & 64.45 & 77.14 & 78.32 \\
\textbf{Top 5 feat. (5)}   & 3213.2 & 36.34 & 66.89 & 67.90 \\
\textbf{Top 4 feat. (4)}   & 2720.1 & 17.47 & 65.40 & 66.10 \\
\textbf{Top 3 feat. (3)}   & 6244.4 &  7.77 & 62.78 & 63.72 \\
\textbf{Top 2 feat. (2)}   & 4912.5 &  2.41 & 59.86 & 60.70 \\
\bottomrule
\end{tabular}
\caption{Comparison of ARF Performance on the Network dataset ~\cite{Xavier2024} with different set of features - 80:20 train/test prequential split - SHAP runtime (including setting the explainer and inference time) for $n_\text{samples} = 100$ - $n_\text{trees} = 10$}
\label{tab:shap-runtime-network}
\end{table}

Using a simple online training setting, Table \ref{tab:shap-runtime} and \ref{tab:shap-runtime-network} report a SHAP runtime growing exponentially as the number of features increases - reaching more than 49 minutes for only 1,000 samples with 8 features for the \emph{Electricity} dataset. Moreover, we observe that incorporating more features can improve accuracy for both datasets, at the cost of the SHAP runtime. 

These trends are further supported by Table \ref{tab:shap-runtime-tree-size} and Figure \ref{fig:runtime-shap-tree-size} in Appendix \ref{sec:shap-runtime-limit}, which illustrates the exponential impact of model size on SHAP runtime. This limitation motivates the need for lightweight, scalable methods for estimating feature importance in ARF models - enabling the identification of the most influential features without compromising runtime or predictive performance. 

\noindent In comparison, MDI is significantly less computationally expensive than SHAP as it only relies on the prediction paths within the model and so avoids SHAP's exhaustive search of all feature combinations\footnote{Approximate versions of SHAP have been proposed that avoid the complete search of feature combinations. However, theses methods are not applicable in dynamic streaming scenarios where the feature set may vary through time.}. However, MDI is sensitive to the performance of the model and underperforms during learning phases \cite{molnar2018interpretable,rajbahadur2022ImpactofFeatureImportance}. Consequently, neither SHAP or MDI in their standard configuration are ideal for streaming data scenarios.

\newpage

\subsection{Impact of the number of models on SHAP Runtime}\label{sec:shap-runtime-limit}

\begin{figure}[h]
    \centering
    \includegraphics[width=0.95\linewidth]{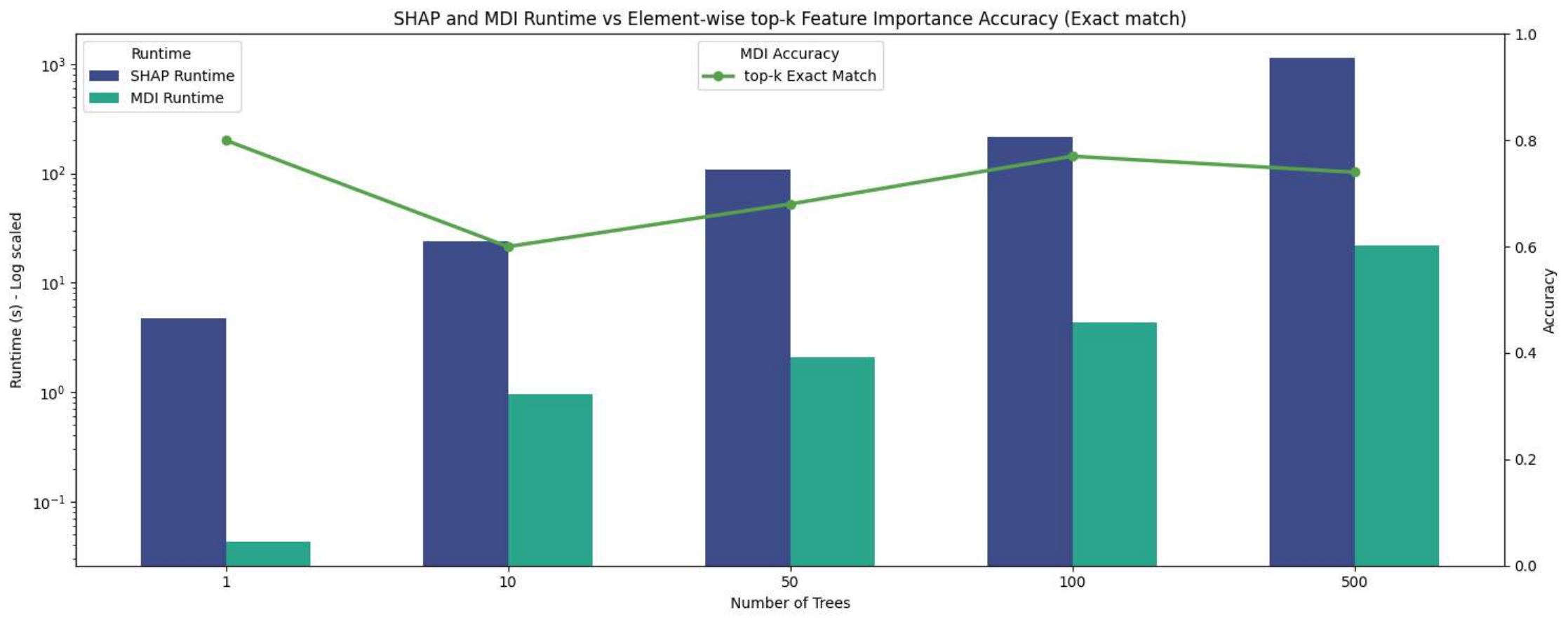}
    \caption{Influence of number of trees in the forest on Model and Feature importance algorithms Performance - Synthetic dataset with 80:20 train/test split - Metrics of algorithms are given for $n_\text{samples} = 100$ samples from the test dataset}
    \label{fig:runtime-shap-tree-size}
\end{figure}

Using the SEA synthetic dataset\footnote{\url{https://riverml.xyz/0.14.0/api/datasets/synth/SEA/}}, Table \ref{tab:shap-runtime-tree-size} compares runtime and the Top-k exact match accuracy of element-wise feature importance of the MDI compared to SHAP. Figure \ref{fig:runtime-shap-tree-size} illustrates the results shown in the Table.

\begin{table}[h]
\centering
\footnotesize
\begin{tabular}{>{\centering\arraybackslash}p{1.5cm}>{\centering\arraybackslash}p{1.7cm}>
{\centering\arraybackslash}p{1.6cm}>
{\centering\arraybackslash}p{1.6cm}>{\centering\arraybackslash}p{1.5cm}>{\centering\arraybackslash}p{1.6cm}>{\centering\arraybackslash}p{1.6cm}>{\centering\arraybackslash}p{1.7cm}}
\toprule
\textbf{ \# Trees} & \textbf{Training Runtime} & \textbf{Training Accuracy} & \textbf{Testing Accuracy} & \textbf{SHAP Runtime} & \textbf{MDI Runtime} & \textbf{Top-k exact} \\
\midrule
1 & 0.1794 & 0.89 & 0.95 & 4.7189 & 0.0431 & 0.80 \\
10 & 1.4716 & 0.95 & 0.98 & 23.6637 & 0.9536 & 0.60 \\
50 & 7.5501 & 0.96 & 0.98 & 108.6549 & 2.0682 & 0.68 \\
100 & 14.8777 & 0.96 & 0.99 & 216.1023 & 4.3310 & 0.77 \\
500 & 79.0376 & 0.96 & 0.99 & 1129.4253 & 22.0117 & 0.74 \\
\bottomrule
\end{tabular}
\caption{Influence of number of trees in the forest on Model and Feature importance algorithms Performance - Synthetic dataset with 80:20 train/test split - Metrics of algorithms are given for $n_\text{samples} = 100$ samples from the test dataset}
\label{tab:shap-runtime-tree-size}
\end{table}

\newpage

\subsection{KS-TEST} \label{sec:ks_test_theory}

\textbf{Kolmogorov-Smirnov Test}. The Kolmogorov-Smirnov test ~\cite{KS_test} is a metric used to quantify the distance between two data distributions. Given two batches of streaming data, each containing $n$ samples and $d$ features, the test computes the maximum difference between their empirical cumulative distribution functions. Specifically, for two batches of streaming data, $ X_{t-1} = \{ x^{t-1}_{1}, x^{t-1}_{2}, \dots, x^{t-1}_{d} \} $ and $ X_{t} = \{ x^{t}_{1}, x^{t}_{2}, \dots, x^{t}_{d} \} $, the KS statistic is defined as:
\begin{equation}\label{eq:ks_test}
D_i = \sup_{x_i} \left| F_{t}(x^{t}_i) - F_{t-1}(x^{t-1}_i) \right|
\end{equation}
where \( F_{t-1}(x_i) \) and \( F_{t}(x_i) \) are respectively the empirical cumulative distribution functions of the previous and current batches for the feature $i$. This statistic quantifies the largest deviation between the distributions of the two batches. A higher value of $D_i$ suggests a more significant shift in the feature's distribution, indicating potential drift. While other methods like ADWIN \cite{bifet2007ADWIN} are tailored for continuous streaming scenarios with adaptive windowing, the KS-Test is better suited for our batch-wise sequential training setup (see Section \ref{sec:experimental-step}).

\subsection{DAFI algorithm} \label{sec:appendix-DAFI-algorithm}

\textbf{Mathematical formalization.} Recalling expression of SHAP, MDI and KS-Test (Equations \ref{eq:shap}, \ref{eq:mdi} and \ref{eq:ks_test}), we can mathematically formalize the DAFI algorithm. For any sample $\mathbf{x} = \{x_1, \dots, x_d\}$, we express $\Phi_j$, the feature importance of $x_j$ computed by DAFI:
\begin{equation}
\Phi_j = \mathbf{1}_{\text{drift}}(\mathbf{x}) \cdot \frac{\left| \phi_j \right|}{\sum_{i \in \{1, \ldots, d\}} \left| \phi_i \right|} + \left(1 - \mathbf{1}_{\text{drift}}(\mathbf{x})\right) \cdot \frac{\varphi_j}{\sum_{i \in \{1, \ldots, d\}} \varphi_i}
\end{equation}
\[
\text{with } 1_{\text{drift}}(\mathbf{x}) = \begin{cases}
1 & \text{if } \left( \sum_{i \in \{1, ..., d\}} D_i \right) > \eta \\
0 & \text{otherwise}
\end{cases}
\]
where $\Phi_j$, $\phi_j$, and $\varphi_j$ are respectively the FI of the feature j in $\mathbf{x}$ computed by DAFI, SHAP and MDI algorithms. $1_{\text{drift}}$ represents the indicator function of the KS-Test drift detector, which returns 1 if the distributional shift between the current and previous batches exceeds a predefined threshold $\eta$, and 0 otherwise. 

\textbf{Algorithm.} Building on the preceding definitions, we lists the DAFI procedure below (Algorithm \ref{alg:drift_aware_feature_importance})

\begin{algorithm}[h]
\caption{DAFI - Drift-aware Feature Importance algorithm}\label{alg:drift_aware_feature_importance}
\footnotesize
\begin{algorithmic}[1]

\Require{ARF model $\mathcal{T} = \{T_1, T_2, \dots, T_m\}$,  $\mathbf{x} = \{x_1, \dots, x_n\}$, Drift\_threshold $\eta$}

\vspace{0.5em}

\State Detect drift and warning/drift from ARF:
\State $\text{Drift\_Detected} \gets \textbf{False}$
\State $\text{Drift\_Accum} \gets \textbf{0}$ \Comment{\textit{Set a variable to accumulate the drifts observed for each features}}

\vspace{0.5em}

\For{each feature $x_i \in \mathbf{x}$}
    \State $\text{Drift\_Accum} \gets \text{KS\_STATS}(\text{batch\_train[i-1]}, \text{batch\_train[i]})[x_i]$ 
    
    \If{$\text{Drift\_Accum} > \eta$}
        \State $\text{Drift\_Detected} \gets \textbf{True}$
        \State \textbf{break} \Comment{\textit{Stop if drift is detected for any feature}}
    \EndIf
\EndFor

\vspace{0.5em}

\If{\text{Drift\_Detected}} 
    \State $\text{importance} \gets \text{SHAP}(\mathbf{x})$

\vspace{0.5em}

\Else 
    \State $\text{importance} \gets \text{MDI}(\mathbf{x})$
\EndIf

\State \Return $\text{Normalized Importance}$ for all $x_i \in \mathcal{x}$

\end{algorithmic}
\end{algorithm}

\newpage

\subsection{Dataset description} \label{sec:dataset-description}

Table \ref{tab:info-dataset} describes each dataset.

\begin{table}[h]
\centering
\footnotesize
\begin{tabular}{p{1.2cm}>{\centering\arraybackslash}p{1.5cm}>{\centering\arraybackslash}p{1.2cm}>
{\centering\arraybackslash}p{2cm}>{\centering\arraybackslash}p{2cm}>{\centering\arraybackslash}p{1.7cm}>{\centering\arraybackslash}p{1.5cm}}
\toprule
\textbf{Datasets} & \textbf{Ref.} & \textbf{Domain} & \textbf{\# of Samples} & \textbf{\# of Features} & \textbf{Target} & \textbf{\# of Class} \\
\midrule
Electricity & ~\cite{gomes_adaptive_2017} & Energy & 45,312 & 8 & Class & 2 \\
Weather & ~\cite{kaggle_weather} & Climate & 145,460 & 23 & RainTomorrow & 2 \\
Network & ~\cite{Xavier2024} & 6G & 3,175,140 & 46 & mob\_pattern & 2 \\
\bottomrule
\end{tabular}
\caption{Information about selected datasets}
\label{tab:info-dataset}
\end{table}

The Electricity dataset contains electricity price, ask and demand of the Australian New South Wales Electricity Market sampled at 5 minute intervals. The dataset was introduced by ~\cite{harries1999splice2}, refined by ~\cite{Gama2004}, and later used by ~\cite{gomes_adaptive_2017}, to predict whether the moving average of the price of the last 24 hours goes UP or DOWN. Features selected for this dataset are: ``period'', ``transfer'', ``nswprice'', ``vicprice'', ``vicdemand'', and ``nswdemand''. Electricity markets are inherently dynamic, influenced by fluctuation in supply, demand, and external events as weather or policy changes. As a result, models trained on historical electricity data must be adaptable to changing patterns and shifts in market distribution.

The Weather dataset includes daily observations of various Australian meteorological stations. Used by ~\cite{Raksha2021}, and available from ~\cite{kaggle_weather}, it contains various numerical and categorical data from sensors, such as temperature, humidity, wind speed, pressure, and rainfall. Features selected for this dataset are: ``MinTemp'', ``WindGustSpeed'', ``Evaporation'', ``MaxTemp'', ``Rainfall'', and ``Sunshine''. The objective of this dataset is to predict whether it will rain or not the following day. Weather prediction is useful for a wide range of applications, including agriculture, disaster preparedness, transportation, and outdoor event planning. However, weather data is subject to seasonal variations, sensor malfunction, and evolving patterns, making it prone to data distribution shifts over time. Consequently, drift detection techniques can help maintain predictive accuracy by adapting models to changing environmental conditions. 

The Network dataset consists of Radio Access Network (RAN) Key Performance Indicator (KPI) and Performance Monitoring (PM) data collected from the OpenIreland testbed (~\cite{XAVIER2024110710}, \cite{Xavier2024}). The dataset spans various traffic classes, categorized as either “Benign” (e.g., web browsing, VoIP, IoT, and YouTube video streaming) or “Attack” (e.g., DDoS Ripper, DoS Hulk, PortScan, and Slowloris). In a real-world deployment, mobile networks often consist of a mix of multivendor equipment, where certain measurements may be available in some network nodes but missing in others. Additionally, legacy hardware may only support a subset of parameters, limiting direct comparisons across different network elements. As a result, models trained on such data must be robust to evolving parameters and changing feature availability. This variability underscores the need for drift detection techniques to ensure models remain accurate over time despite shifts in feature distributions. Furthermore, explainability in AI is critical for diagnosing performance fluctuations and understanding why certain classifications are made, particularly when key network metrics are missing or when unexpected behavior arises due to hardware or configuration differences. The data reflects mobility patterns, including static, pedestrian, car, bus, and train scenarios. The dataset features detailed measurements derived from the MAC and PHY layers of the network. At the PHY layer, metrics such as Channel Quality Indicator (CQI), Modulation and Coding Scheme (MCS), and Signal-to-Interference-plus-Noise Ratio (SINR) for both PUSCH and PUCCH channels are included. These metrics quantify channel quality, signal interference, and modulation efficiency. The MAC layer data focuses on traffic behavior and service interactions, including metrics such as data rate, successfully delivered packets, dropped packets, and buffer status. The dataset is useful for analyzing real-world mobile communication environments and classifying RAN mobility types while remaining independent of specific user equipment behavior. Additionally, with both benign and attack traffic classes, it supports the development and validation of intrusion detection systems to enhance network security strategies. Features selected for this dataset are: ``mac\_dl\_mcs'', ``phy\_ul\_pusch\_sinr'', ``phy\_ul\_pucch\_sinr'', ``phy\_ul\_pusch\_rssi'', ``mac\_dl\_cqi'', and ``phy\_ul\_pucch\_rssi''.

\newpage

\subsection{Illustration Top-k set and match metrics} \label{sec:appendix-topk-illustration}

\begin{figure}[h]
    \centering
    \includegraphics[width=1\linewidth]{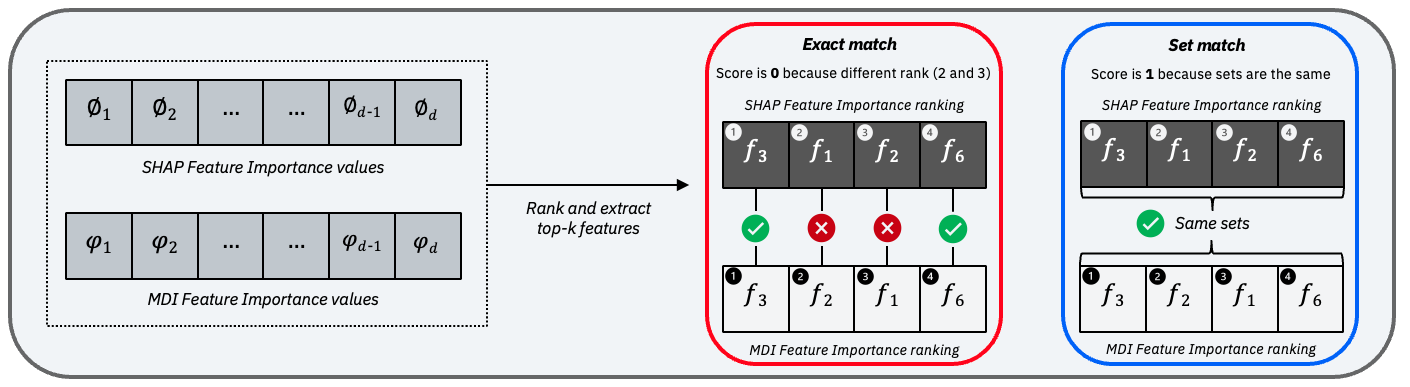}
    \caption{Illustration of the top-k metrics - Exact and Set matches}
    \label{fig:illustration-topk}
\end{figure}

Figure \ref{fig:illustration-topk} illustrates our Top-$k$ feature importance metrics.

\subsection{Results on other datasets} \label{sec:appendix-evaluation-other-datasets}

Table \ref{fig:tab-element-wise-other-datasets} presents the results of our DAFI algorithm on the Electricity and Weather datasets.

\begin{table}[h]
\centering
\footnotesize
\begin{tabular}{p{1.4cm}>{\centering\arraybackslash}p{2.0cm}>
{\centering\arraybackslash}p{1.6cm}>{\centering\arraybackslash}p{1.8cm}>{\centering\arraybackslash}p{1.4cm}>{\centering\arraybackslash}p{1.6cm}>{\centering\arraybackslash}p{1.8cm}}
\toprule
\textbf{Datasets} & \textbf{FI Method} & \textbf{Runtime (s)} & \textbf{Saved runtime (\%)} & \textbf{Top-k set} & \textbf{Top-k exact} & \textbf{Top-k Spearman} \\
\midrule
\multirow{3}{*}{Electricity} & SHAP & 1790.79 & 0.00 & 1.00 & 1.00 & 1.00 \\
& MDI & 6.09 & 99.65 & 0.56 & 0.39 & 0.67 \\ 
& \textbf{DAFI} & 1161.60 & 35.13 & 0.85 & 0.76 & 0.85 \\ \midrule
\multirow{3}{*}{Weather} & SHAP & 1544.23 & 0.00 & 1.00 & 1.00 & 1.00 \\
& MDI & 7.44 & 99.52 & 0.50 & 0.20 & 0.61 \\ 
& \textbf{DAFI} & 1023.09 & 33.75 & 0.82 & 0.66 & 0.83 \\ 
\bottomrule
\end{tabular}
\caption{Runtime and metrics of FI methods applied Element-wise}
\label{fig:tab-element-wise-other-datasets}
\end{table}

\newpage


\end{document}